\documentclass{article}

\usepackage{arxiv}
\usepackage{CJK}
\usepackage[T1]{fontenc}    
\usepackage{hyperref}       
\usepackage{url}            
\usepackage{booktabs}       
\usepackage{amsfonts}       
\usepackage{nicefrac}       
\usepackage{color}
\usepackage{microtype}      
\usepackage{lipsum}
\usepackage{multirow}
\usepackage{graphicx}
\usepackage{amsmath}
\usepackage{amssymb}
\usepackage{makecell}
\DeclareMathOperator*{\argmax}{arg\,max}

\title{Effidit: Your AI Writing Assistant}
\author{
  Shuming Shi, Enbo Zhao, Duyu Tang, Yan Wang, Piji Li, Wei Bi, Haiyun Jiang,\\ \bf{Guoping Huang, Leyang Cui, Xinting Huang, Cong Zhou, Yong Dai, Dongyang Ma}\\
  \\
  Tencent AI Lab\\
  ailabnlp@tencent.com
}

\begin{document}
\begin{CJK*}{UTF8}{gbsn}
\maketitle

\begin{abstract}
In this technical report, we introduce Effidit (\textbf{Eff}icient and \textbf{I}ntelligent E\textbf{dit}ing), a digital writing assistant that facilitates users to write higher-quality text more efficiently by using artificial intelligence (AI) technologies.
Previous writing assistants typically provide the function of error checking (to detect and correct spelling and grammatical errors) and limited text-rewriting functionality.
With the emergence of large-scale neural language models, some systems support automatically completing a sentence or a paragraph.
In Effidit, we significantly expand the capacities of a writing assistant by providing functions in five categories: text completion, error checking, text polishing, keywords to sentences (K2S), and cloud input methods (cloud IME).
In the text completion category, Effidit supports generation-based sentence completion, retrieval-based sentence completion, and phrase completion. In contrast, many other writing assistants so far only provide one or two of the three functions.
For text polishing, we have three functions: (context-aware) phrase polishing, sentence paraphrasing, and sentence expansion, whereas many other writing assistants often support one or two functions in this category.

The main contents of this report include major modules of Effidit, methods for implementing these modules, and evaluation results of some key methods.
\end{abstract}

\section{Introduction}
A digital writing assistant is a software program that uses artificial intelligence (AI) technology to help users with their writing process.
Typical writing assistants often provide functions like error checking and sentence rewriting. Error checking is about the automatic detection and correction of spelling and grammatical errors, which could alleviate skilled writers from tedious proofreading work and help unskilled editors and non-native speakers reduce errors in their writing. Sentence rewriting is about generating new sentences (hopefully with improved quality or preferred styles) that have similar meaning as the original one.
Thanks to large-scale neural language models like GPT-3 ~\cite{brown2020language}, now some writing assistants can complete a sentence, a paragraph, or even an article, according to user-provided text as the prefix.

In spite of the nice features provided by state-of-the-art writing assistants \cite{Grammarly, QuillBot, Wordtune, BaiduCreation, WPS, AimWriting, MetaXiezuoCat, Mypitaya}, they are still far from satisfactory, in terms of the scope and quality of their functions.

In this technical report, we introduce Effidit (/i$'$fidit/, \textbf{Eff}icient and \textbf{I}ntelligent E\textbf{dit}ing)\footnote{https://effidit.qq.com/en}, a system that significantly expands the capacities of a typical writing assistant by providing five categories of functions: text completion, error checking, text polishing, keywords to sentences (K2S), and cloud input methods (cloud IME). For most of these functions, we support both Chinese and English.

Major functions of Effidit are summarized below,

\begin{itemize}
\item \textbf{Text completion}: In this category, Effidit supports generation-based sentence completion, retrieval-based sentence completion, and phrase completion, among which only one or two functions are supported by most other writing assistants. It is worth noting that the phrase completion of Effidit is not necessarily prefix-based, as illustrated by the left example of Figure-~\ref{fig:phrase_completion_example}.

\item \textbf{Error checking}: For error checking on English text, Effidit offers similar functions to other systems. For Chinese error correction, among the three error types of word substitution, insertion, and deletion, the performance of Effidit is in the first tier in correcting word substitution errors and far ahead in handling insertion and deletion errors.

\item \textbf{Text polishing}: Three functions are supported: (context-aware) phrase polishing, sentence rewriting, and sentence expansion. In contrast, many other writing assistants often support one or two of the three functions in this category.

\item \textbf{K2S}: The K2S module of Effidit takes one or more keywords as input and returns a list of sentences. Each output sentence either contains the input keywords, or is semantically related to them. The K2S function is built for the scenario that users only have some basic concepts in their minds but do not know how to organize sentences to express their ideas.

\item \textbf{Cloud IME}: Effidit provides a Chinese Pinyin input method as well as an English IME. Instant suggestions pop up when the user is typing in our text editor. With the help of the large-scale language models hosted on our servers, Effidit has the potential to give richer suggestions than the input methods installed on the local desktop.
\end{itemize}

This report is organized as follows: In Section~\ref{sec: toolkits}, we take a detailed look at the overall system and its major functions, from the viewpoint of users. Then we discuss, in Section ~\ref{sec: implementation}, the technology and implementation of some key modules. Finally we give a summary of our system and directions for future improvement.

\begin{figure}[htp]
    \centering
    \includegraphics[width=16cm]{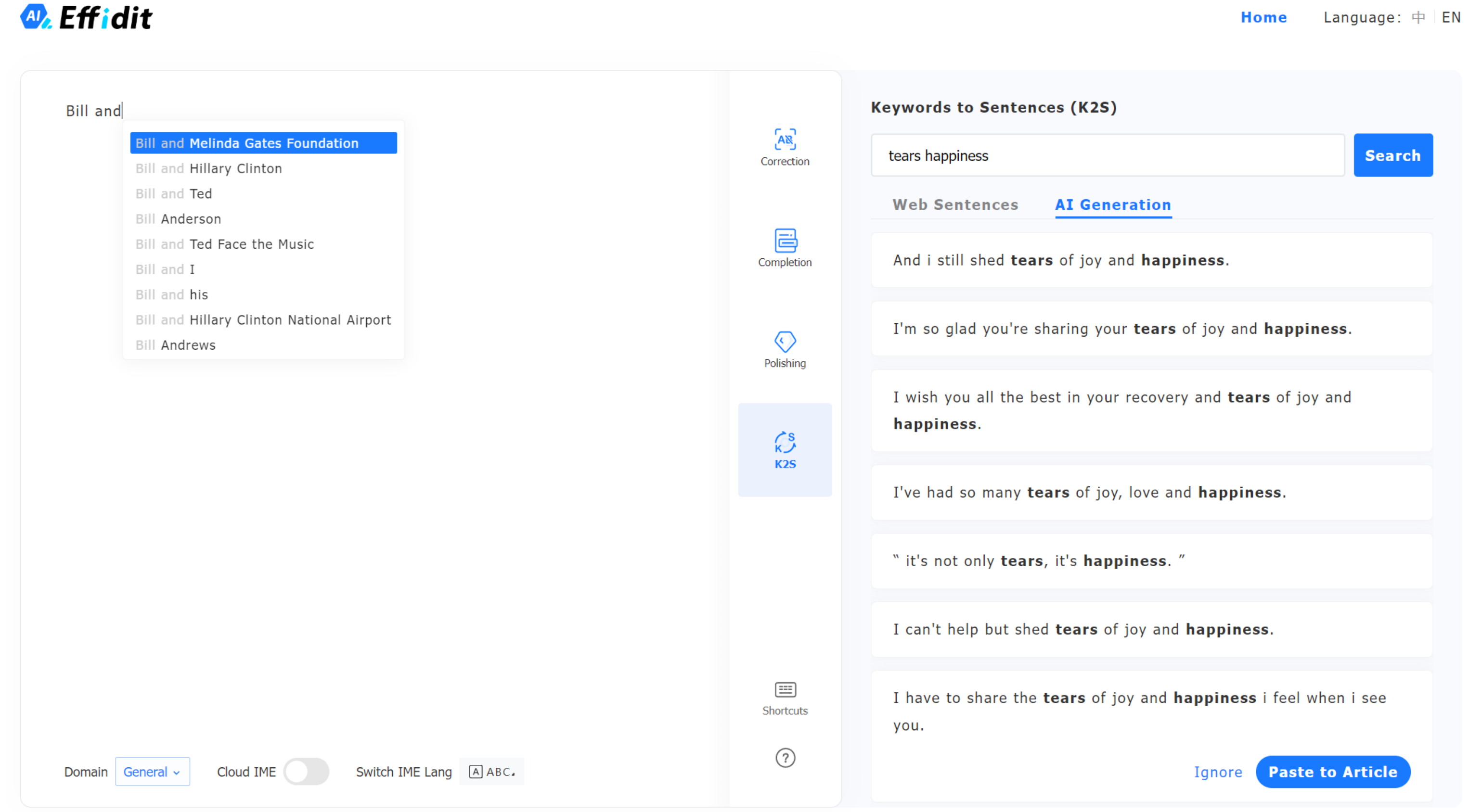}
    \caption{Screenshot of the Effidit online demo}
    \label{fig:system_overview}
\end{figure}

\section{System Overview}
\label{sec: toolkits}
The first version of Effidit was released in April, 2022.\footnote{(in Chinese) https://mp.weixin.qq.com/s/b-kPSR3aFPKHpUnFv7gmeA} Users can experience the system by accessing the online demo\footnote{ https://effidit.qq.com/demo\_en} or downloading the desktop client.
Figure~\ref{fig:system_overview} shows a screenshot of the online demo of Effidit.
The left part is a plain-text editor for users to add and edit text, whereas the right pane is for triggering most core functions of Effidit and displaying corresponding results. In addition, there are some UI elements at the bottom for changing domains and setting up the cloud IME.

Two domains are supported so far\footnote{More domains will be added in future versions.}: the \textbf{general domain} for general text editing, and the \textbf{academic domain} for facilitating the writing of academic documents.
Five types of functions are offered in the general domain: text completion, error checking, text polishing, keywords to sentences (K2S), and cloud IME. These functions are described in the following subsections.
In addition to the general functions, additional features are provided for the academic domain, as being illustrated in Section~\ref{sec:academic_domain}

\subsection{Text Completion}
\label{sec:text_completion_intro}

Effidit provides both phrase-level and sentence-level text completion suggestions.
These suggestions may allow users to quickly complete their sentences or provide ideas for how to proceed with the writing process.

\begin{figure}[htp]
    \centering
    \includegraphics[width=8cm]{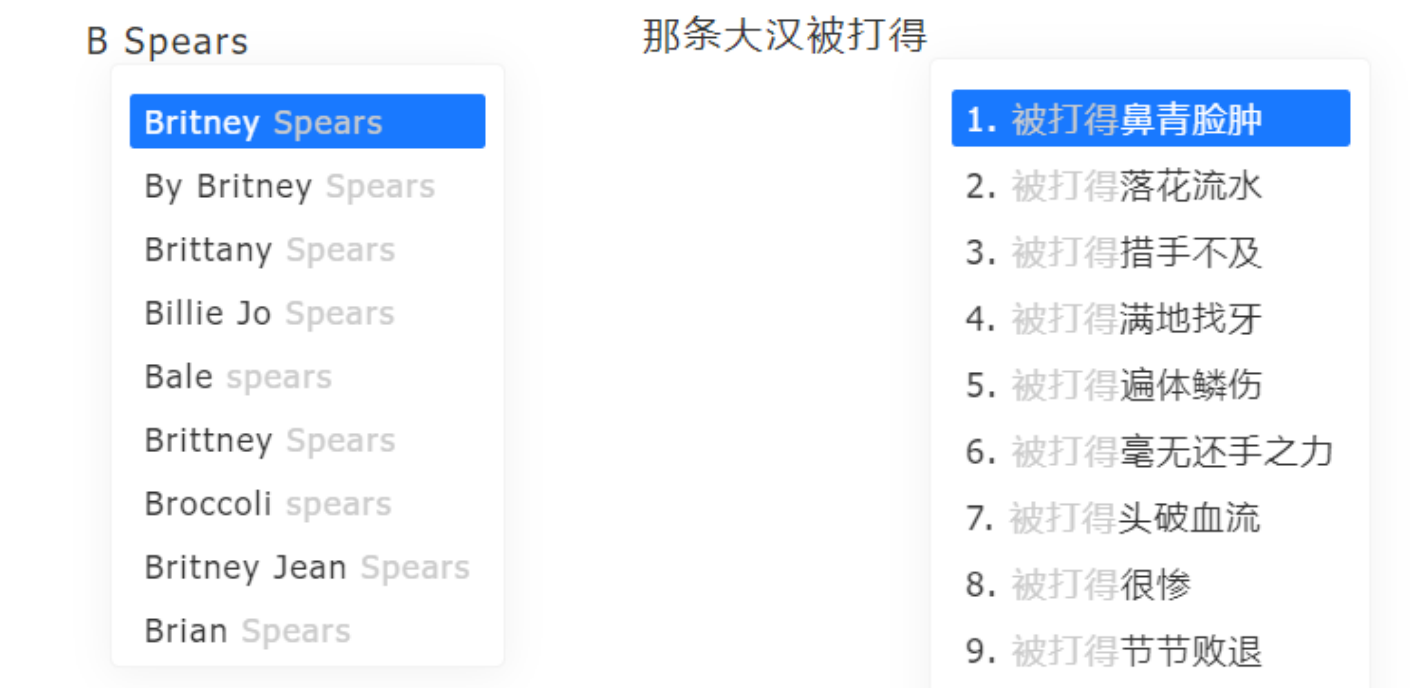}
    \caption{Phrase completion examples: An English (left) and a Chinese (right) example}
    \label{fig:phrase_completion_example}
\end{figure}

\textbf{Phrase completion}: Figure~\ref{fig:phrase_completion_example} shows two examples of phrase completion, one for English and one for Chinese.
Please note that Effidit considers both prefix and suffix in generating phrase completion candidates, whereas most other writing assistants only consider prefix information.
As presented in the left part of the figure, when the caret is after the first letter ``B'', the top completion results contain both the prefix ``B'' and the suffix ``Spears''.
High quality phrase completion may help to not only improve writing efficiency, but reduce the chance of errors. For instance, when a user types ``Los A'' and triggers phrase completion, ``Los Angeles'' appears as the first result. In this case, common spelling mistakes like ``Los Angelas'' can be avoided.
In both the online demo and the desktop client, the keyboard shortcut for phrase completion is [Ctrl]+[;] (i.e., CTRL plus the key on the right of the L key).

\begin{figure}[htp]
    \centering
    \includegraphics[width=16cm]{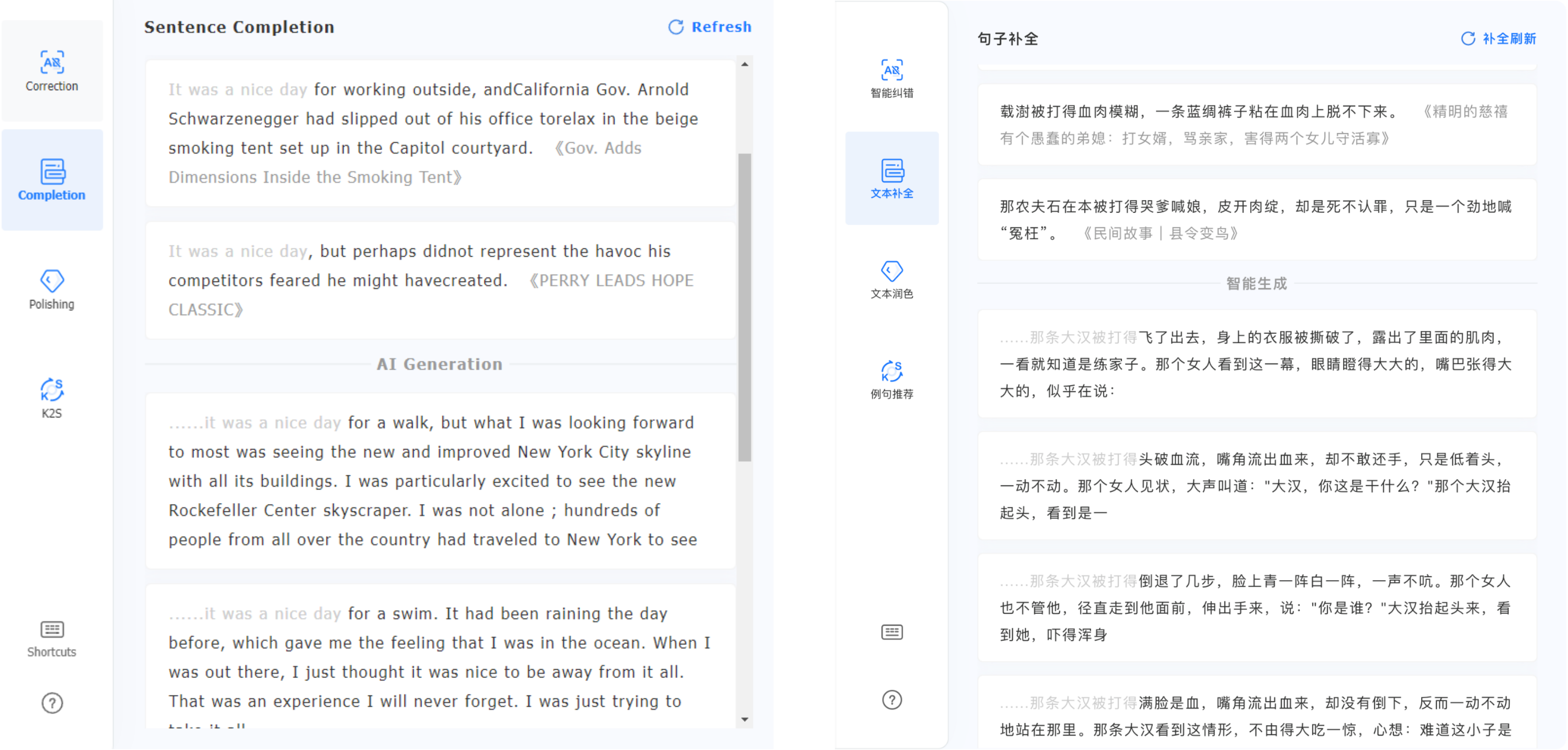}
    \caption{Examples of sentence-level completion. The upper two are example sentences from the Web, and the others are generation-based completion suggestions}
    \label{fig:sentence_completion_example}
\end{figure}

\textbf{Sentence completion}: Given a prefix, Effidit returns two types of sentence completion suggestions: Web sentences, and AI generation.
Figure~\ref{fig:sentence_completion_example} demonstrates both types of completion results for an English prefix (``It was a nice day'') and a Chinese prefix (``那条大汉被打得''), respectively.
Web sentences are the results of a retrieval-based method, where top sentences that match the prefix well are retrieved from a corpus of high-quality sentences collected beforehand from the Web.
AI generation is based on large-scale neural language models (refer to Section~\ref{sec:Text_completion_details} for technical details).
For both types of completion suggestions, Effidit displays multiple candidates, from which the user can select the most appropriate one to adopt (for AI generation) or for reference (for both Web sentences and AI generation).
The keyboard shortcut for triggering sentence completion is [Ctrl]+['] (i.e., CTRL plus the second key on the right of the L key).

By phrase completion, retrieval-based sentence completion, and generation-based sentence completion, Effidit provides comprehensive and multi-dimensional text completion suggestions for enhancing writing efficiency, reducing errors, or inspiring ideas.

\subsection{Text Polishing}

Writers often need to modify or rewrite their sentences for readability and clarity, to avoid self-repetition, or to make the text more attractive. Rewriting sentences manually can be tedious or inefficient. Writing assistants could help alleviate the tedious aspects of text rewriting and improve writing efficiency by providing high-quality rewriting candidates to choose from.
Effidit provides a variety of text polishing functions, including phrase polishing, sentence paraphrasing, and sentence expansion.

\begin{figure}[htp]
    \centering
    \includegraphics[width=11cm]{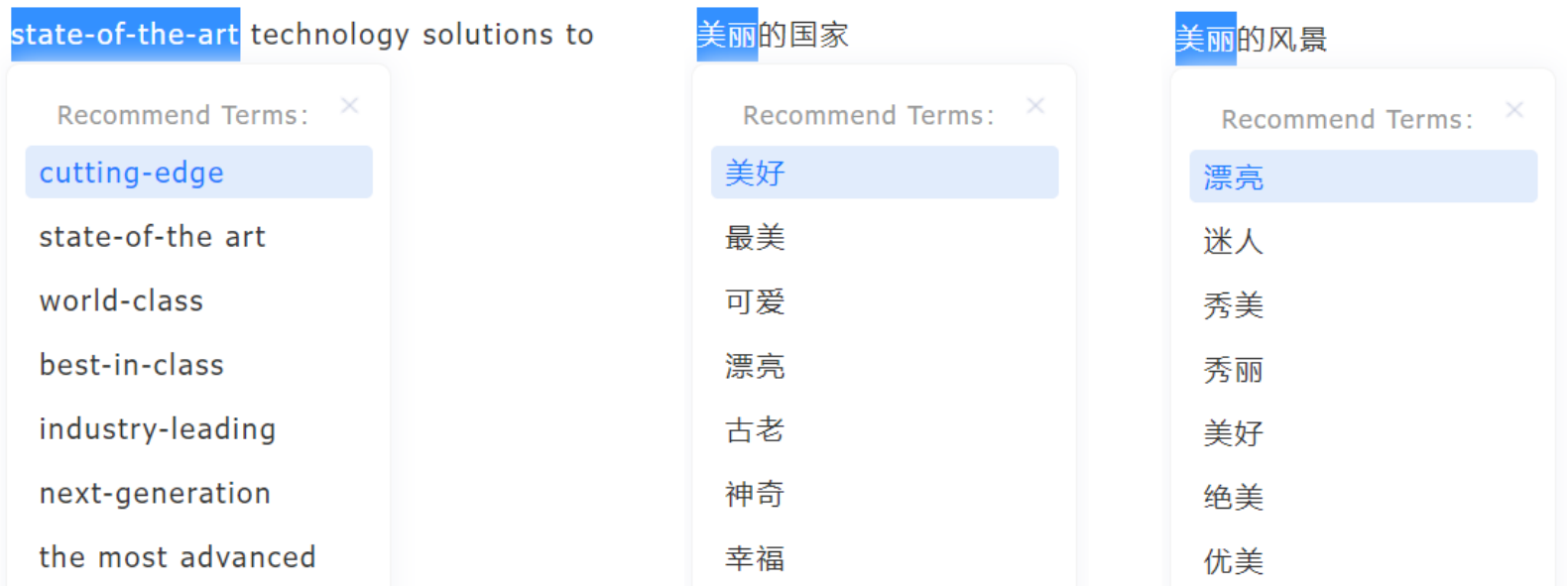}
    \caption{Examples of phrase polishing}
    \label{fig:phrase_polishing_example}
\end{figure}

In \textbf{phrase polishing}, users can select a word or phrase in a sentence and ask Effidit to suggest word/phrase candidates that have similar meaning to the selected text piece and also fit the current context. Figure~\ref{fig:phrase_polishing_example} shows a few examples where an English phrase (``state-of-the-art'') and a Chinese phrase (``美丽'') are selected for triggering the phrase polishing function.
Compared with most other writing assistants, the phrase polishing of Effidit has two major advantages: 1) The polishing approach is context-aware, but not a simple synonym replacement. 2) The triggering rate of Effidit is higher than that of most similar systems. In other words, Effidit is able to generate suggestions for more words or phrases.

\begin{figure}[htp]
    \centering
    \includegraphics[width=8cm]{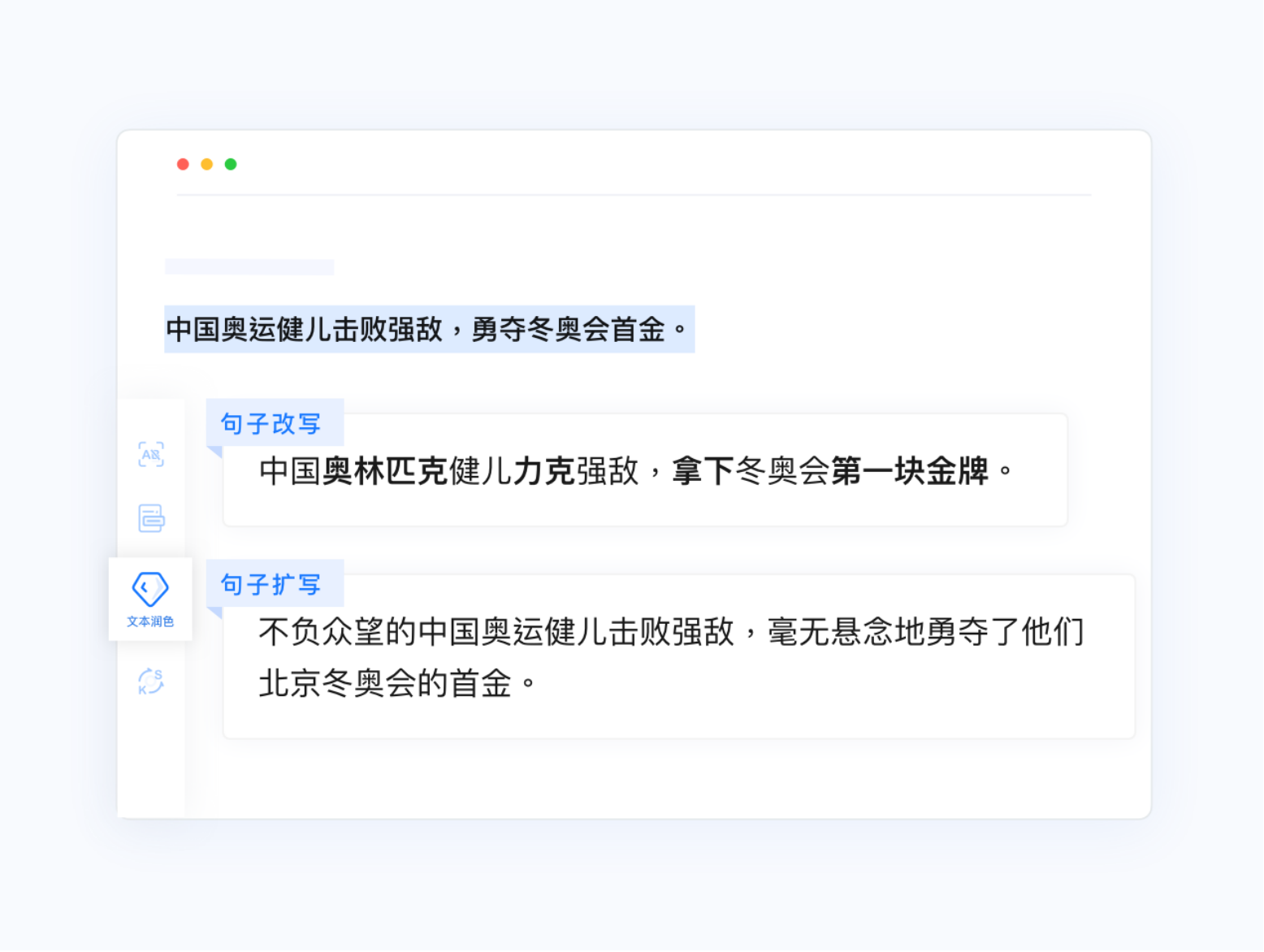}
    \caption{An example of Chinese sentence paraphrasing and expansion}
    \label{fig:sentence_polishing_example}
\end{figure}

\textbf{Sentence paraphrasing} is the task of rewriting an input sentence in different wording while maintaining its semantic meaning. \textbf{Sentence expansion} is the process of adding modifiers to an input sentence to make a longer sentence with rich information.
The paraphrasing and expansion of Chinese sentences have been implemented in Effidit and made online. An example is shown in Figure~\ref{fig:sentence_polishing_example}.
The English version of the two functions is in progress.

The keyboard shortcut for text polishing is [Ctrl]+[,] (i.e., CTRL plus the key on the right of the M key). If the selected text piece is a phrase, phrase polishing results are shown; otherwise sentence paraphrasing and expansion are triggered.

\subsection{Error Correction}
\begin{figure}[htp]
    \centering
    \includegraphics[width=16cm]{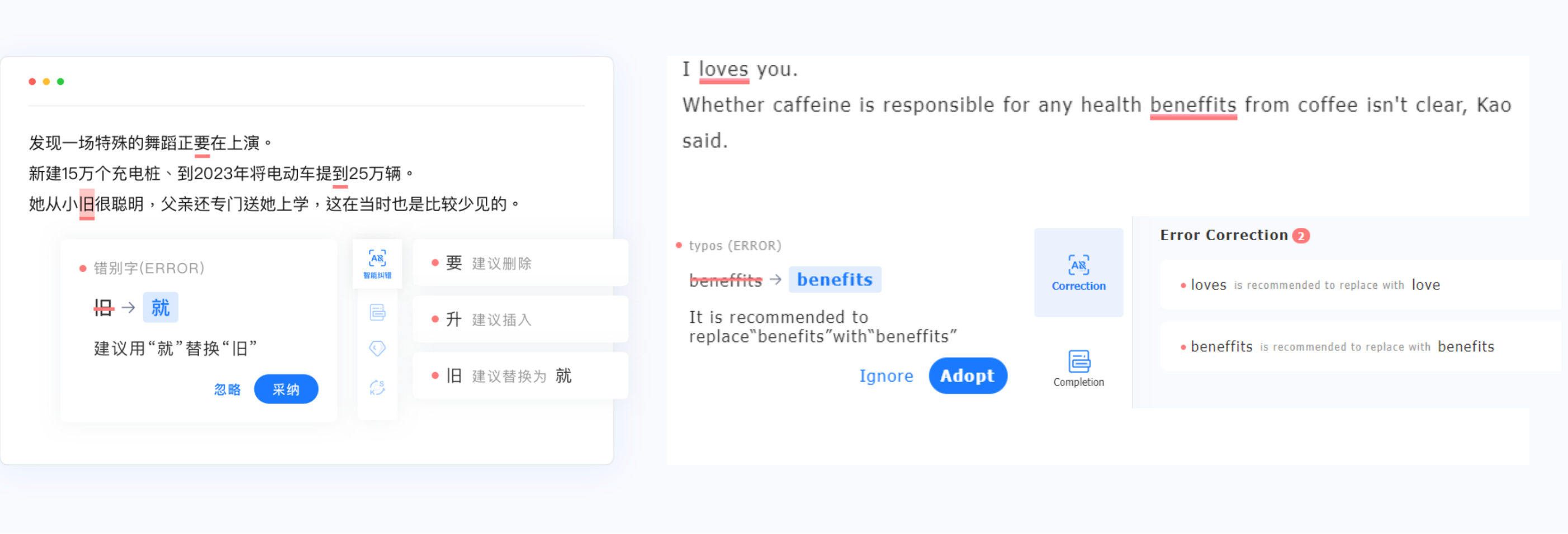}
    \caption{Examples of error detection and correction}
    \label{fig:ErrorDetection}
\end{figure}

Error detection and correction is a must-have feature for writing assistants. Major error types for Chinese text include substitution errors (replacing a word with another one), insertion errors (inserting a missing word) and deletion errors (deleting a redundant word).
Effidit supports high-quality Chinese error detection and correction, especially on the insertion and deletion types of errors.
Figure~\ref{fig:ErrorDetection} displays several examples that Effidit is able to handle correctly.
Effidit also supports error correction for English text, as shown in the figure.
The keyboard shortcut for error correction is [Ctrl]+[M].

\subsection{K2S}
\begin{figure}[htp]
    \centering
    \includegraphics[width=16cm]{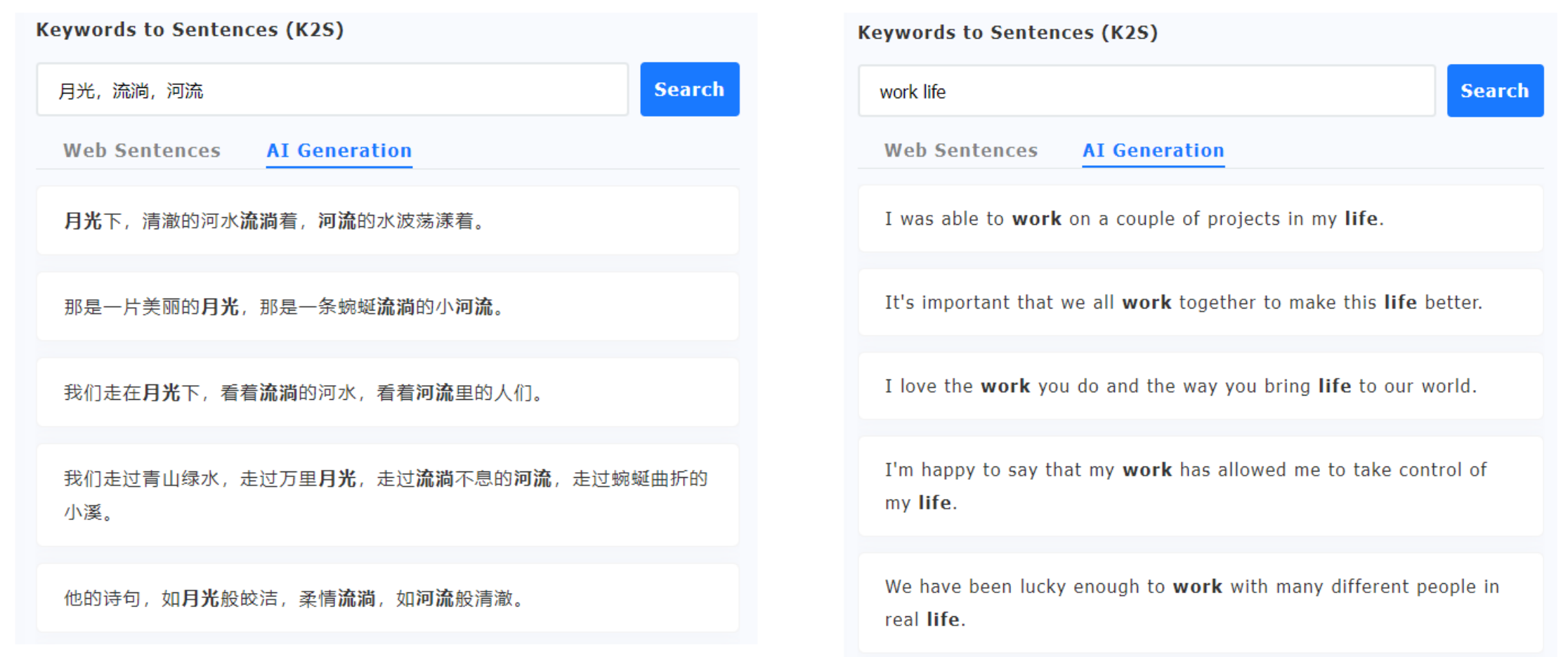}
    \caption{Examples of K2S}
    \label{fig:k2s}
\end{figure}

The K2S module of Effidit takes one or more keywords as input and returns a list of sentences. Each output sentence either contains the input keywords, or is semantically related to them. The K2S function is built for the scenario that users only have some basic concepts in their minds but do not know how to organize sentences to express their ideas.
Effidit returns two types of sentences for the input keywords: Web sentences, and AI generation. Figure~\ref{fig:k2s} demonstrates both types of results for Chinese keywords ``月光\;流淌\;河流'' and English keywords ``work life'' respectively.
Web sentences are the results of a retrieval-based method, where top sentences containing the input keywords are retrieved from a corpus of high quality sentences collected beforehand from the Web.
AI generation is based on large-scale neural language models (refer to Section~\ref{sec:K2S_detail} for technical details).
For both types of results, Effidit displays multiple candidates, from which the user can select the most appropriate one to adopt (for AI generation) or for reference (for both Web sentences and AI generation).

\subsection{Cloud IME}
\begin{figure}[htp]
    \centering
    \includegraphics[width=7cm]{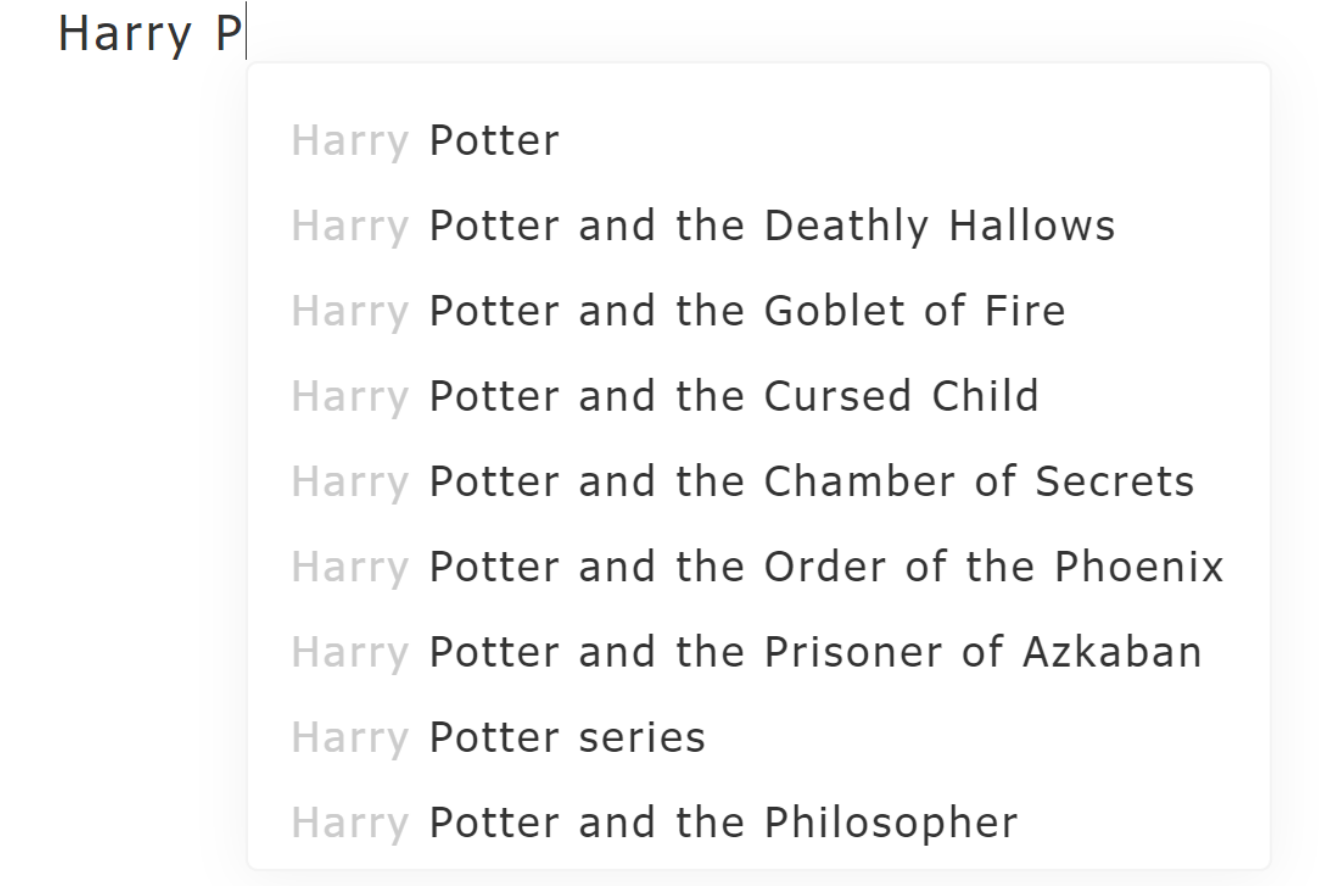}
    \caption{An example of cloud IME}
    \label{fig:CloudIME}
\end{figure}

Effidit provides cloud input methods for both Chinese and English to improve the input efficiency of words and phrases.
Instant suggestions pop up when the user is typing in our text editor. Figure~\ref{fig:CloudIME} depicts an example where the names of some novels in the Harry Potter series are shown as suggestions.
With the help of the large-scale language models hosted on our servers, Effidit is able to generate long phrases or medium-frequency ones as suggestion results, whereas the input methods installed on the local desktop typically suggest single words or high-frequency phrases. Effidit also has the potential to generate more appropriate suggestions based on larger context.

\subsection{Academic Domain}
\label{sec:academic_domain}
Effidit supports two domains so far: the general domain, and the academic domain.
In addition to the above functions for helping with general text editing, two additional features are provided in the academic domain to facilitate the writing of academic documents.

\begin{figure}[htp]
    \centering
    \includegraphics[width=16cm]{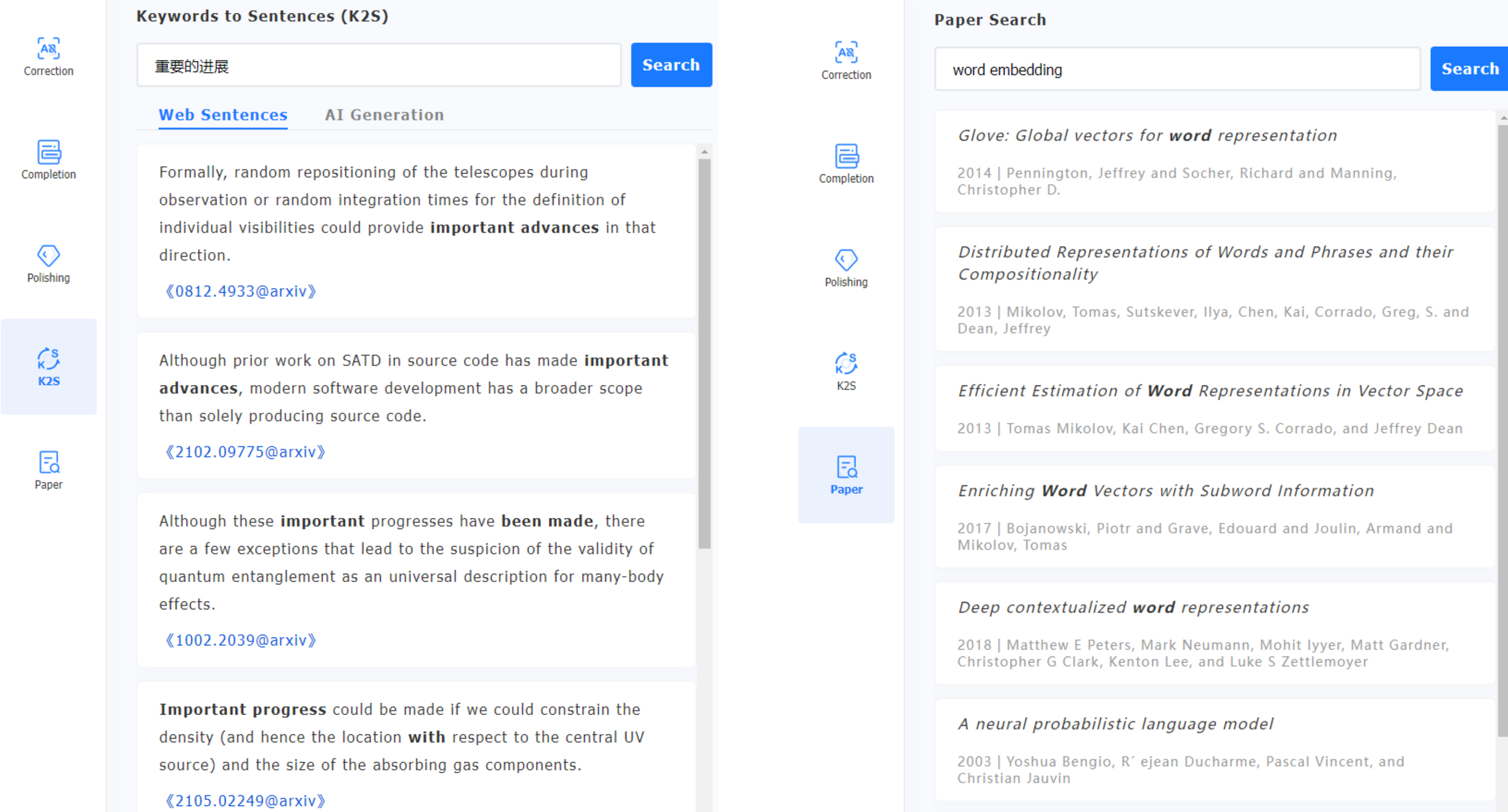}
    \caption{Left: Cross-lingual K2S, Right: Semantic-enhanced paper search}
    \label{fig:academic}
\end{figure}

\textbf{Cross-Lingual K2S}: Please recall that the K2S module takes one or more keywords as input and returns a list of sentences. Effidit supports cross-lingual K2S in the academic domain, where the output is English sentences and the input can be a mixture of Chinese and English keywords. Please refer to the left part of Figure~\ref{fig:academic} for an example, where the input Chinese keywords mean ``important progress''.
The output sentences either illustrate the translation of the keywords in English, or demonstrate the usage of the input keywords in academic papers.
The goal is to facilitate non-native English speakers in writing academic documents.
As related work, ESODA \cite{ESODA} is another system that supports cross-lingual retrieval of academic sentences.

\textbf{Semantic-Enhanced Paper Search}: Unlike other paper search services such as Google Scholar \cite{GoogleScholar}, the paper search of Effidit focuses more on recalling papers that are semantically related to the input keywords other than those containing the input keywords in their titles or contents. In particular, when the input keyword is a research topic, important papers on this topic will be returned even if some keywords are missing in some results. For example, as shown in the right part of Figure~\ref{fig:academic}, for query ``\textit{word embedding}'', not all top results contain the word ``\textit{embedding}'' in their titles or contents.

\section{Implementation}
\label{sec: implementation}

\subsection{Generation-Based Sentence Completion}
\label{sec:Text_completion_details}
\begin{figure}[h] 
  \centering    
  \setlength{\abovecaptionskip}{3pt}
  \includegraphics[width=0.8\textwidth]{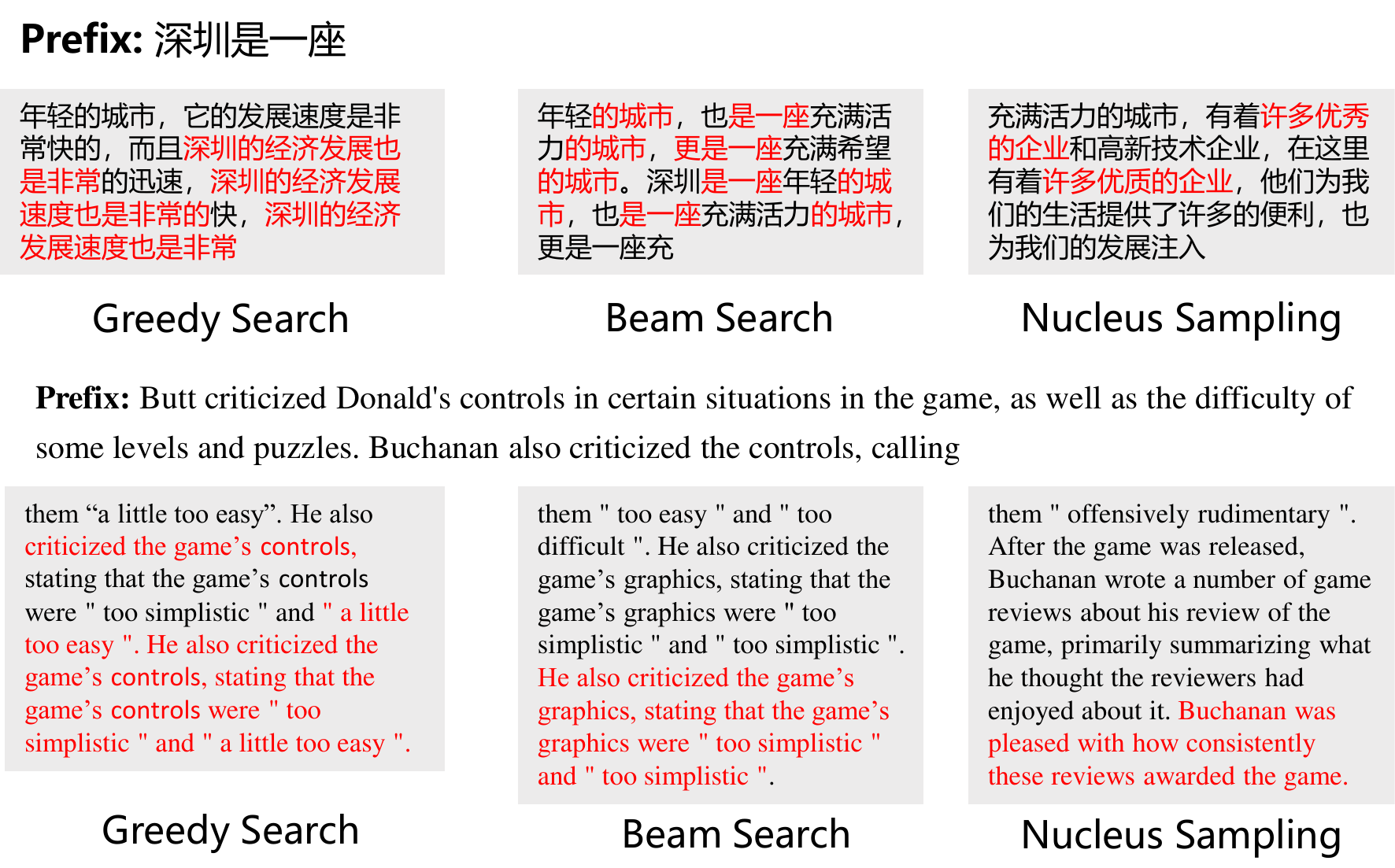}
  \caption{The degeneration problems in GPT-2}
  \label{fig:degeneration_example}
  \vspace{-1.5mm}
\end{figure}

In this subsection\footnote{Please note that most of the techniques in this section have been released. One may refer to our arXiv paper ``A Contrastive Framework for Neural Text Generation'' for more technical details and experimental results \cite{su2022contrastive}.}, we introduce the training and decoding methods of our generation-based sentence completion model.
One conventional approach is training a language model with a large corpus and decoding the most likely token as the next token.
However, since these models are usually trained with Maximum Likelihood Estimation (MLE), they may lead to a serious degeneration problem, i.e., the generated texts from the language model tend to be dull and contain undesirable repetitions at different levels (e.g., token-, phrase-, and sentence-level)~\cite{DBLP:journals/corr/abs-1902-00098}. Some examples of the degeneration problem are shown in Figure~\ref{fig:degeneration_example}, where we can find many repetitions.

We argue that this degeneration stems from the \textit{anisotropic} distribution of token representations, i.e., their representations reside in a narrow subset of the entire space~\cite{DBLP:conf/emnlp/Ethayarajh19,DBLP:conf/icml/DongCL21,su2021tacl}.
In Effidit, we implement \textit{SimCTG} (a \underline{\textbf{sim}}ple \underline{\textbf{c}}ontrastive framework for neural \underline{\textbf{t}}ext \underline{\textbf{g}}eneration)~\cite{su2022contrastive}, which encourages the model to learn discriminative and isotropic token representations.
A novel decoding strategy, \textit{contrastive search}, is also employed to complement SimCTG.
Key intuitions behind contrastive search are: (i) At each decoding step, the output should be selected from the set of most probable candidates predicted by the model to better maintain the semantic coherence between the generated text and the human-written prefix. (ii) The sparseness of the token similarity matrix of the generated text should be preserved to avoid degeneration.
For more details about SimCTG and its potential extensions, please refer to ~\cite{su2022contrastive, su2021tacl, su2022MAGIC}.

\subsubsection{Methodology}
In this section, we first present how to apply contrastive learning to calibrate the representation space of the language model. Then, we introduce the contrastive search decoding algorithm.

\paragraph{Contrastive Training}
\label{sec:contrastive_training}
Our goal is to encourage the language model to learn discriminative and isotropic token representations. To this end, a contrastive objective $\mathcal{L}_{\textup{CL}}$ is introduced into the training of the language model. Specifically, given a variable-length sequence $\boldsymbol{x}=\{x_1, ..., x_{|\boldsymbol{x}|}\}$, $\mathcal{L}_{\textup{CL}}$ is defined as 
\begin{equation}
    \label{eq:cl}
    \mathcal{L}_{\textup{CL}} = \frac{1}{|\boldsymbol{x}|\times(|\boldsymbol{x}| - 1)}\sum_{i=1}^{|\boldsymbol{x}|}\sum_{j=1,j\neq i}^{|\boldsymbol{x}|}\max\{0,\rho - s(h_{x_i}, h_{x_i}) + s(h_{x_i}, h_{x_j})\},
\end{equation}
where $\rho\in[-1,1]$ is a pre-defined margin and $h_{x_i}$ is the representation of token $x_i$ produced by the model. The similarity function $s$ computes the cosine similarity between token representations as
\begin{equation}
    \label{eq:cosine}
    s(h_{x_i}, h_{x_j}) = \frac{h_{x_i}^\top h_{x_j}}{\|h_{x_i}\|\cdot\|h_{x_j}\|}.
\end{equation}
Intuitively, by training with $\mathcal{L}_{\textup{CL}}$, the model learns to pull away the distances between representations of distinct tokens.\footnote{By definition, the cosine similarity $s(h_{x_i}, h_{x_i})$ of the identical token $x_i$ is $1.0$.} Therefore, a discriminative and isotropic model representation space can be obtained. The overall training objective $\mathcal{L}_{\textup{SimCTG}}$ is then defined as
\begin{equation}
    \label{eq:simctg}
    \mathcal{L}_{\textup{SimCTG}} = \mathcal{L}_{\textup{MLE}} + \mathcal{L}_{\textup{CL}},
\end{equation}
where the vanilla maximum likelihood estimation (MLE) objective $\mathcal{L}_{\textup{MLE}}$ is:
\begin{equation}
    \label{eq:mle}
    \mathcal{L}_{\textup{MLE}} = -\frac{1}{|\boldsymbol{x}|}\sum_{i=1}^{|\boldsymbol{x}|}\log p_{\theta}(x_i|\boldsymbol{x}_{<i}).
\end{equation} Note that, when the margin $\rho$ in $\mathcal{L}_{\textup{CL}}$ equals to $0$, the $\mathcal{L}_{\textup{SimCTG}}$ degenerates to the vanilla MLE objective $\mathcal{L}_{\textup{MLE}}$. 

\paragraph{Contrastive Search}
We employ a new decoding method called \textit{contrastive search}. At each decoding step, the key ideas of contrastive search are (i) the generated output should be selected from the set of most probable candidates predicted by the model; and (ii) the generated output should be discriminative enough with respect to the previous context. In this way, the generated text can better maintain semantic coherence with respect to the prefix while avoiding model degeneration.

Formally, given the context $\boldsymbol{x}_{<t}$, at time step $t$, the selection of the output $x_t$ follows
\begin{equation}
    \label{eq:score}
    x_t = \argmax_{v\in V^{(k)}}\bigg\{(1 - \alpha)\times \underbrace{p_{\theta}(v|\boldsymbol{x}_{<t})}_{\textup{model confidence}} -  \: \alpha \times \underbrace{(\max\{s(h_v, h_{x_j}):1\leq j \leq t-1\})}_{\textup{degeneration penalty}}\bigg\},
\end{equation}
where $V^{(k)}$ is the set of $k$ most-probable candidates predicted by the model $\theta$ and $k$ is typically set as 3$\sim$10. In Eq. \ref{eq:score}, the first term, \textit{model confidence}, is the probability of candidate $v$ predicted by the model. The second term, \textit{degeneration penalty}, measures how discriminative of candidate $v$ with respect to the previous context $\boldsymbol{x}_{<t}$ and $s$ is defined in Eq. \ref{eq:cosine}.
Specifically, it is defined as the maximum cosine similarity between the representation of $v$ and that of all tokens in $\boldsymbol{x}_{<t}$. Here, the candidate representation $h_v$ is computed by the model given the concatenation of $\boldsymbol{x}_{<t}$ and $v$.
Intuitively, a larger degeneration penalty of $v$ means it is more similar to the context, therefore more likely to lead to model degeneration.
The hyperparameter $\alpha\in[0,1]$ regulates the importance of the two components. When $\alpha=0$, contrastive search degenerates to the greedy search method.

\subsubsection{Experiments}
\label{sec:experiment}
We evaluate several approaches on the task of open-ended document generation.

\textbf{Model and Baselines.} SimCTG is architecture-agnostic and can be applied to any generation model.
In this work, we evaluate SimCTG on the representative GPT-2 model~\cite{radford2019language}.
Specifically, we fine-tune GPT-2 on the evaluation benchmark (detailed below) with the proposed objective $\mathcal{L}_{\textup{SimCTG}}$ (Eq. \ref{eq:simctg}) and generate the text continuation with different decoding methods.
Experiments are conducted using the base model (117M parameters) which consists of 12 Transformer layers~\cite{DBLP:conf/nips/VaswaniSPUJGKP17} with 12 attention heads.
SimCTG is compared with two strong baselines: (1) GPT-2 fine-tuned with the standard MLE objective (Eq. \ref{eq:mle}); and (2) GPT-2 fine-tuned with the unlikelihood objective ~\cite{DBLP:conf/iclr/WelleckKRDCW20}.\footnote{The unlikelihood baseline is implemented with the official code, which can be found here \url{https://github.com/facebookresearch/unlikelihood_training}.} Our implementation is based on the Huggingface library~\cite{DBLP:journals/corr/abs-1910-03771}.

\textbf{Evaluation Benchmark.} Experiments are conducted on the Wikitext-103 dataset ~\cite{DBLP:conf/iclr/MerityX0S17} which contains a large collection of Wikipedia articles with over 100 million words and 260 thousand unique tokens. Wikitext-103 is a document-level dataset and has been widely used for the evaluation of large-scale language modeling ~\cite{DBLP:conf/acl/DaiYYCLS19,DBLP:conf/iclr/KhandelwalLJZL20,DBLP:journals/tacl/YogatamadK21}.

\begin{table*}[h]
    \small
    \centering  
    \renewcommand{\arraystretch}{1.2}
    \setlength{\tabcolsep}{6pt}
    \scalebox{0.85}{
    \begin{tabular}{ccccc}
        \hline
        \textbf{Model}&Decoding Method&Coherence&Fluency&Informativeness\\
        \hline
        Agreement&-&0.51&0.64&0.70\\
        \hline
        \multirow{2}{*}{MLE}&nucleus&2.92&3.32&3.91\\
        &contrastive&2.78&2.29&2.56\\
        \hline
        \multirow{2}{*}{Unlikelihood}&nucleus&2.59&3.02&3.58\\
        &contrastive&2.76&2.90&3.35\\
        \hline
        \multirow{2}{*}{SimCTG}&nucleus&2.96&3.34&3.96\\
        &contrastive&3.25$^{\bigstar}$&3.57$^{\bigstar}$&3.96\\
        \hline
        \multirow{2}{*}{SimCTG-large}&nucleus&3.01&3.37&\textbf{3.98}\\
        &contrastive&\textbf{3.33}$^{\bigstar}$&\textbf{3.66}$^{\bigstar}$&\textbf{3.98}\\
        \hline
        Human&-&3.70&3.71&4.21\\
        \hline
    \end{tabular}}
    \caption{Human evaluation results. The results labeled with a star (${\bigstar}$) significantly outperform the corresponding nucleus sampling results (Sign test with p-value < 0.05).}
        \vspace{-1.5mm}
    \label{tb:human_evaluation}
\end{table*}

\paragraph{Human Evaluation}
\label{sec:human_evaluation_detail}
Human evaluation is conducted with the help of graders proficient in English from a third-party annotation company. We randomly select 200 prefixes with a length of 32 from the test set of Wikitext-103. For each prefix, we use different models (MLE, Unlikelihood, and SimCTG) with two decoding methods (nucleus sampling and contrastive search) to generate text continuations with length of 128. The evaluation follows a 5-point Likert scale (1, 2, 3, 4, or 5) for each of the following features:
\begin{itemize}
    \item \textbf{Coherence}: Whether the generated text is semantically consistent with the prefix.
    \item \textbf{Fluency}: Whether the generated text is fluent and easy to understand.
    \item \textbf{Informativeness}: Whether the generated text is diverse and contains interesting content. 
\end{itemize}

Table \ref{tb:human_evaluation} presents the human evaluation results, with the first row showing strong inter-annotator agreements as measured by Fleiss$^\prime$ kappa coefficient~\cite{fleiss1971mns}. We observe that the results of SimCTG + contrastive search are significantly better than nucleus sampling with different models in terms of coherence and fluency (sign test with p-value $\textless$ 0.05). In addition, SimCTG-large + contrastive search achieves the best performance across the board and even performs comparably with human-written text on the fluency metric (sign test with p-value $\textgreater$ 0.4).
This reveals the clear generality of our approach to large size models. Future work could focus on extending it to models that contain over billions of parameters such as GPT-3~\cite{brown2020language}.

\begin{table*}[h]
    \small
    \centering  
    \renewcommand{\arraystretch}{1.2}
    \setlength{\tabcolsep}{6pt}
    \scalebox{0.75}{
    \begin{tabular}{ccccc}
        \hline
        \textbf{prefix}&\makecell[l]{Butt criticized Donald's controls in certain situations in the game, as well as the difficulty \\of some levels and puzzles. Buchanan also criticized the controls, calling}&div.&coh.&gen-ppl\\
        \hline
        
        beam&\makecell[l]{them " too easy " and " too difficult ". \textcolor{red}{He also criticized the game's graphics, stating that the game's}\\ \textcolor{red}{graphics were " too simplistic " and " too simplistic ".} He also criticized the game's graphics, stating\\ that the game's graphics were " too simplistic " and " too simplistic ". He also criticized the graphics,\\ stating that the game's graphics were " too simplistic " and " too simplistic ". He also criticized the\\ graphics, stating that the game's graphics were " too simplistic " and " too simplistic ". He also\\ criticized the graphics, stating that the game's graphics were" ...}&0.02&0.605&7.91\\
        \hline
        nucleus&\makecell[l]{them " offensively rudimentary ". After the game was released, Buchanan wrote a number of game\\ reviews about his review of the game, primarily summarizing what \textcolor{blue}{he thought the reviewers had}\\ \textcolor{blue}{enjoyed about it.} \textcolor{blue}{Buchanan was pleased with how consistently these reviews awarded the game.} \\For one, \textcolor{blue}{he praised the puzzles}, compared them to Nos. 1 and 2 and \textcolor{blue}{enjoyed the construction of the}\\ \textcolor{blue}{artificial intelligence in the game's forms}. \textcolor{blue}{He found no fault with the puzzles}, even in those situations\\ where they seemed repetitive. According to an Xbox Magazine review, " every time ...}&0.98&0.597&37.83\\
        \hline
        contrastive&\makecell[l]{them " unimpressive " and a " nightmare " of an experience to play with players unfamiliar with\\ Tetris. On the other hand, his opinion was shared by other reviewers, and some were critical of the\\ game's technical design for the Wii version of Tetris. In addition, Tintin's review included a quote\\ from Roger Ebert, who said that Tetris was better than the original game due to its simplicity and\\ ease of play. Ebert's comments were included in the game's DVD commentary, released on March\\ 22, 2010. It is unclear if any of the video commentary was taken from ...}&0.98&0.626&19.64\\
        \hline
    \end{tabular}}
    \caption{\textbf{Case study}: The beam search produces degeneration repetitions (highlighted in red) and the nucleus sampling produces text that has incoherent semantics with respect to the prefix (highlighted in blue). The ``div.'' and ``coh.'' stand for diversity and coherence metrics. (best viewed in color)}
    \label{tb:case_study}
\end{table*}

\paragraph{Case Study}
Table \ref{tb:case_study} shows some examples generated by SimCTG with different decoding methods given a specific prefix. We can see from the results that beam search produces undesirable sequence-level repetitions, resulting in low diversity and low generation perplexity.
On the other hand, in the prefix, the person ``Buchanan'' \textit{criticizes} the game.
However, the results from nucleus sampling contain contradicting statements, resulting in a low coherence score as well as a high generation perplexity. Lastly, for contrastive search, it generates text that is semantically consistent to the prefix with a proper perplexity while obtaining the same diversity as that of the nucleus sampling.

\subsection{Grammatical Error Correction}
\label{sec:Grammatical_Correction}
Given a piece of written text as the input, the task of grammatical error correction is to detect spelling and grammatical errors and give correction suggestions. 
Grammarly \cite{omelianchuk2020gector} achieves remarkable performance in handling grammatical errors in English text.
In this part, we describe our system for Chinese error correction.
Dominating error types in Chinese text include substitution, deletion, and insertion errors. 
We have a tail-to-tail model \cite{DBLP:conf/acl/Li020} to handle substitution errors and a tailored pretrained model \cite{zhouinsertbert2022} to handle deletion and insertion errors.

\subsubsection{Tail-to-Tail Non-Autoregressive Sequence Prediction}
Our first model is a non-autoregressive sequence-to-sequence model, abbreviated as \textbf{TtT}~\cite{DBLP:conf/acl/Li020}. We use the BERT encoder to produce the contextual vector representation for each token.
A Conditional Random Fields (CRF) \cite{DBLP:conf/icml/LaffertyMP01} layer is stacked on the up tail to model the dependencies among neighbouring tokens. The focal loss penalty strategy \cite{DBLP:journals/pami/LinGGHD20} is adopted to alleviate the class imbalance problem considering that most of the tokens in a sentence are not changed.

\subsubsection{Pretraining for Detecting Word Insertion and Deletion Errors}
Our second model is specifically designed to tackle insertion and deletion errors.
Recently, BERT and its model variations \cite{hong2019faspell,zhang-etal-2020-spelling,liu-etal-2021-plome} show promising results on handling grammatical errors related to word substitution. 
However, they do not perform well in handling word insertion and deletion errors.
The reason is that the learning objective of BERT assumes the existence of a word at each position, but this is incapable of determining whether no word exists at a position. 
Take word insertion as an example. If a word needs to be inserted between two words (i.e., $w_i$ and $w_{i+1}$), the standard BERT hardly detects anomalies because $w_i$ fits well into its preceding context and $w_{i+1}$ fits well into its following context. 

\begin{figure}[h]
	\centering
	\includegraphics[width=\textwidth]{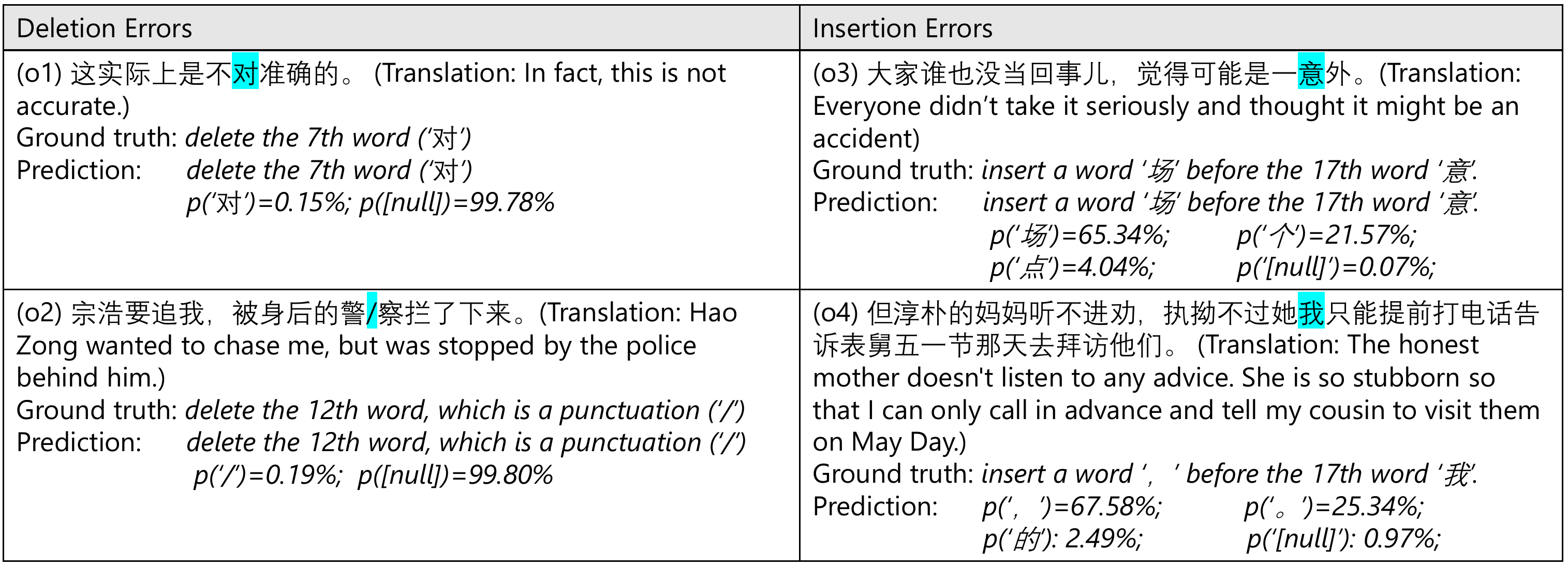}
	\caption{Model outputs of InsertBERT for word deletion and insertion errors. Erroneous positions are highlighted in blue. }
	\label{fig:insertbert-case-study}
\end{figure}

We address the aforementioned problem with a tailored pretrained model named InsertBERT \cite{zhouinsertbert2022}.
The basic idea is to enable the model to determine whether a word exists at a particular position.
We achieve this by introducing a special token \texttt{[null]}, the prediction of which stands for the non-existence of a word.
In the training stage, we design pretraining tasks such that the model learns to predict \texttt{[null]} and real words jointly given the surrounding context. 
In the inference stage, the model readily detects whether a word should be inserted or deleted in the standard masked language modeling manner.
An additional advantage of the model is that it conducts detection and correction simultaneously because the top ranked real word can be directly used for error correction.

Figure \ref{fig:insertbert-case-study} shows the model outputs of InsertBERT for both deletion and insertion errors. We can see that InsertBERT is capable of handling both normal words (o1 and o3) and punctuations (o2 and o4).

\subsection{Text Polishing}
\subsubsection{Phrase Polishing}

Given a word or phrase in a sentence, the phrase polishing module suggests a list of word/phrase candidates that have similar semantic meaning to the input word/phrase and also fit the current context well.
Phrase polishing is fundamentally the context-aware semantic expansion task introduced in \cite{han2020case}, where a neural method was introduced to address the problem.
Since the neural method in that paper is quite time-consuming, a light-weight alternative approach is proposed in TexSmart \cite{zhang2020texsmart} to implement the semantic expansion of entities. We make an adaptation to this light-weight approach to support general words and phrases other than just entities.

Our phrase polishing approach includes an offline phase and an online phase.
In the offline phase, we build a similarity graph of words and phrases, where the similarity score is calculated by combining word embedding, distributional similarity, and pattern-based methods \cite{mikolov2013distributed, song2018directional, shi2010corpus}.
The graph is organized in a way that the top neighbors of a node can be retrieved efficiently.

In the online phase, the top neighbors of the input word/phrase are first retrieved from the graph as candidate results. Then a score is calculated for each candidate by considering two factors: the semantic similarity between the candidate and the input word/phrase, and the coherence between the candidate and the context of the input word/phrase in the sentence. Finally, the candidates are sorted by scores and the best scoring candidates are chosen as the phrase polishing results.

Now the core challenge is how to calculate the score of a candidate. We choose to compute the score as a linear combination of $S_1$ and $S_2$, where $S_1$ is the similarity between the candidate and the input word or phrase. Score $S_1$ can be obtained directly from the similarity graph, which has been built offline.

Score $S_2$ is adopted as an estimation of how well the candidate fits the context of the input word/phrase in the sentence.
Suppose $T$ is a candidate, and context $\mathbf{C}=\{c_1, c_2, \cdots, c_k\}$ is the set of words and phrases appearing in a window of the input word/phrase within the sentence. Score $S_2$ is calculated as:
\begin{equation}
  \textrm{$S_2$}(\mathbf{T}, \mathbf{C}) = \frac{1}{k}\sum_{x\in \mathbf{C}} \cos(v_x, w_T)
  \label{eq:s2}
\end{equation}
\noindent where $v_x$ denotes the input embedding of $x$, $w_y$ is the output word embedding of $T$ from a well-trained word embedding model, and $\cos$ is the cosine similarity function.

\subsubsection{Sentence Paraphrasing}

Sentence paraphrasing is about rewriting a sentence with different words or syntactic structures while keeping the same semantics of the original sentence \cite{bhagat2013paraphrase,su2021keep}.
Rewriting provides users with new expressions and insights, which may inspire users to write more elegant text.

The biggest challenge in implementing text rewriting is how to acquire large-scale high-quality paraphrase pairs for training rewriting models.
In Effidit, we adopt three strategies to obtain the training data from various sources.

(1) Back translation\cite{federmann2019multilingual,mallinson2017paraphrasing}: We translate a sentence $S$ into sentence $T_1$ of another language and then translate it back to $T_2$ of the original language. Then the ($S$, $T_2$) pair is added to our training corpus of parallel text.

(2) Retrieval: Given a sentence, we retrieve the most similar sentences from a large-scale corpus using embedding-based retrieval technologies, where SimCSE \cite{gao2021simcse} is adopted as the text encoder and Faiss (Facebook AI Similarity Search) \cite{johnson2019billion} as the retrieval framework.

(3) Collecting existing text matching and paraphrase datasets \cite{liu2018lcqmc,chen2018bq}.

Some sentence pairs obtained from the above methods may not be good paraphrases.
So we delete the pairs with small lexical distances, by calculating their Levenshtein distance and WMD \cite{kusner2015word}.
We also remove the pairs with small semantic similarity, by training some text matching models, e.g., SimCSE \cite{gao2021simcse}.
We model text rewriting as a seq2seq learning task, and train a BART model \cite{lewis2019bart} from the sentence pairs after cleaning.
At inference time, the BART model takes the input sentence as input and outputs a list of paraphrase candidates.

\begin{table}[!hpb]
    \centering
    \small
    \begin{tabular}{c|l}

      Input   &  作文课上，刘老师带我们做了一个游戏。 \\ \\
      \multirow{2}{*}{Generated Output}   & 1. 本周的作文课上，和蔼的刘老师手把手带我们做了一个有趣的游戏。
      \\ & 2. 那天的作文课上，新来的刘老师兴致勃勃地带我们开心地做了一个益智游戏。
\\\hline
      Input   & 2020年7月4日，林丹发文宣布正式退出国家队。  \\ \\
      \multirow{2}{*}{Generated Output}   & 1. 2020年7月4日，中国名将林丹发文低调地宣布正式无限期退出羽毛球国家队。
      \\ & 2. 2020年7月4日，奥运冠军林丹在微博发文宣布从即日起正式退出曾经所在的国家队。

\\\hline
      Input   & 聂海胜谈中国航天员的未来：会刷新更多纪录。  \\ \\
      \multirow{2}{*}{Generated Output}   & 1. 宇航员聂海胜谈关于中国航天员的未来：太空探索会不断地刷新更多世界纪录。
      \\ & 2. 发言人聂海胜谈中国航天员的未来：载人航天会一次次的刷新更多史无前例的纪录。

    \end{tabular}
    \vspace{3mm}
    \caption{Examples of sentence expansion results generated by Effidit}
    \label{tab:expansion_example}
\end{table}

\subsubsection{Text Expansion}
Sentence expansion is about adding modifiers to an input sentence to make a longer one with rich information.
So far, only the Chinese version of text expansion is implemented in Effidit.
Some examples are shown in Table \ref{tab:expansion_example}.

Two approaches are employed in Effidit to generate expansion results: global expansion and local expansion.

\paragraph{Global Expansion}
In global expansion, a seq2seq model is trained from a large collection of (sentence, expanded-sentence) pairs.
Since these kinds of large-scale sentence pairs are not readily available, we choose to build pairs from sentences.
Given a sentence $S$, we first perform syntactic analysis (using TexSmart \cite{zhang2020texsmart}) to get its syntactic tree representation.
Then a skeleton $T$ of this sentence is obtained by randomly removing some modifiers from the syntactic tree. The $(T, S)$ pair is then treated as a (sentence, expanded-sentence) pair and added to our training data.
For example, given a sentence ``\textit{red is a powerful color that attracts more attention than probably any other color}''\footnote{Please note that, although we take an English sentence as an example here to illustrate our method, our current implementation is on Chinese text.}, if we extract ``\textit{red is a color}'' as one skeleton from the syntactic tree of the original sentence, the following pair can be obtained as one item in the training data:

\begin{center}
(``\textit{red is a color}'', ``\textit{red is a powerful color that attracts more attention than probably any other color}'')
\end{center}

A GPT-2 model \cite{gpt2} is then trained for text expansion based on our training data.
To fit the format of GPT-2, we concatenate the sentences in each pair with a special token ``[SEP]''. For example, if the pair is $(T, S)$, the input to GPT-2 during training is ``T [SEP] S [CLS]'', where ``[CLS]'' is an end indicator.
During inference, given sentence $T$, we use ``T [SEP]'' as the input to the GPT-2 model, and ask it to predict the expansion results.

\paragraph{Local Expansion}
In global expansion, the words or phrases to be enriched are decided by the seq2seq model itself and the decision principle is in a black box.
The goal of local expansion is to have more control about where to perform expansion.
The basic idea is to first choose one or a few places (mainly around a noun or verb) explicitly by a simple algorithm and then use a neural language model to predict some words or phrases that can be added to each place.
We call the two steps \textbf{space selection} and \textbf{modifier prediction}, respectively.
In our current implementation, space selection is based on the part-of-speech (POS) information of words and phrases, whereas modifier prediction is performed by the T5 pre-training model \cite{t5}.
Take the input sentence ``\textit{she saw flowers on the grass}'' as an example. Let's assume that two places are chosen in the space selection step: The first is before the word ``flower'', and the second is after the word ``grass''.
By inserting the ``[MASK]'' special token into the original sentence, we get,

\begin{center}
``\textit{she saw [MASK] flowers on the grass [MASK]}''
\end{center}

The above text is then sent to the T5 model for mask prediction.
Possible prediction results for the first mask might include ``\textit{colorful}'' , ``\textit{lovely}'', ``\textit{some beautiful}'', and the like. 
For the second mask, possible prediction results may include ``\textit{like natural rugs}'', ``\textit{in beautiful sunlight}'', etc.

\subsection{Keywords-to-Sentences (K2S)}
\label{sec:K2S_detail}

We formulate the task of generation-based K2S as a text infilling problem, which is to generate missing spans of text.
Our system considers a general text infilling setting, where the incomplete sentence can contain an arbitrary number of blanks to be filled in, and each blank can involve an unknown number of tokens. Illustrated examples are provided in Table~\ref{tab:infill_example}.
\begin{table}[hpbp]
    \centering
    \small
    \begin{tabular}{c|l}
      Input (keywords)   &  春节,喜庆,团聚\\
      \multirow{2}{*}{Output (sentence)}   & \textbf{春节}是中华民族的传统节日，每年春节，家家户户都会举办一场\textbf{喜庆}的年俗活动，\\
      &来表达对新年的祝福和\textbf{团聚}
\\\hline
      Input (keywords) & rich, money, happy \\
      \multirow{2}{*}{Output (sentence)} & People who are \textbf{rich} and have \textbf{money} are more likely to be \textbf{happy} than those \\
      & who are poor.
\\\hline
      Input (keywords) & we observed, our model, promising results\\
      \multirow{2}{*}{Output (sentence)} & In this section, \textbf{we observed} that the architecture of \textbf{our model} neural network \\
      & is very \textbf{promising}.
    \end{tabular}
    \vspace{3mm}
    \caption{Example inputs and the generated infilled results}
    \label{tab:infill_example}
\end{table}

We implement our text infilling model by language modeling~\cite{Donahue2020EnablingLM}. An example for model training is constructed by masking random text segments of the raw data to generate an incomplete sentence as the input sequence with each mask segment replaced by a special token ``[blank]'', and its masked segments are concatenated as the output sequence with the special token ``[ans]'', as illustrated in Table~\ref{tab:data_example}. We then concatenate the input and output sequence with another special token (i.e. Input [seq] Output) and train an off-the-shelf language model. 

With a converged language model for inference, it takes the incomplete sentence as input and continues the input sentence until the number of generated special token ``[ans]'' equals to that of ``[blank]'' in the input sequence. We can obtain the complete sentence by filling the missing parts in the input sequence with the text spans between special tokens in the output sequence, sequentially.
\begin{table}[htbp]
    \centering
    \small
    \begin{tabular}{c|l}
      Data   & 今年春节到处喜庆洋洋，这是家家户户团聚的节日啊\\
      Input   & [blank]春节[blank]喜庆[blank]团聚[blank] \\
      Output & 今年[ans]到处[ans]洋洋，这是家家户户[ans]的节日啊[ans] \\ \hline
      Data & although they did not have a lot of money she says that she was never so happy. \\
      Input & [blank] money [blank] happy [blank]\\
      Output & although they did not have a lot of [ans] she says that she was never so [ans] . [ans]\\
    \end{tabular}
    \vspace{3mm}
    \caption{Constructed inputs and outputs for training language models}
    \label{tab:data_example}
\end{table}

\paragraph{Experiments}
Here, we train a GPT-2 with 12 Transformer layers from scratch with the 3 billion/3 billion  data constructed from about 1.5 billion /1.3 billion Chinese/English sentences. 
We randomly hold out 1000 test data for automatic and human evaluation.
Automatic evaluation metrics in our evaluation include Distinct-1/2~\cite{li2016diversity} (which calculates the ratio of unique 1/2 grams over all tokens in the test set) and Novelty (which calculates the ratio of added tokens over all tokens per test sample).
We also measure the Quality and Diversity by human annotation. The Quality contains two aspects: (1)~whether the result contains all given keywords and the order of keywords is consistent with the given one; (2)~language fluency. The Diversity measures how much degree the semantic of the infilled sentence is expanded from the incomplete sentence parts.
Five human annotators from a commercial annotating company evaluated the results of different methods using ratings from 1 to 5, where 5 is the best rating for Quality and Diversity. 

We compare our text infilling model by language modeling (LM) with other two  baselines: (i) Retrieval: Building an index for the training corpus and retrieving top sentences for the input keywords. BM25~\cite{robertson2009probabilistic} is adopted as the ranking function during the retrieval process. (ii) Masked language model by BERT~\cite{kenton2019bert}: We use the BERT model to predict a single token for each ``[blank]'' in the incomplete sentence. The results in Table~\ref{tab:text_infill_tab1} and Table~\ref{tab:text_infill_tab2} show that LM achieves the best performance on both sentence quality and diversity.

\begin{table}[htbp]
    \centering
    \begin{tabular}{lrrrrrr}
    \toprule
          & Dist-1 & Dist-2 & Novelty & Quality & Diversity \\\midrule
        LM  &0.76&0.92&5.42&3.57&4.32\\
        Retrieval & 0.93 &0.86&1.99&2.67&2.49\\
        BERT &0.95&0.89&2.10&2.13&2.36 \\
        \bottomrule
    \end{tabular}
    \vspace{3mm}
    \caption{Evaluation results for generation-based K2S on Chinese sentences}
    \label{tab:text_infill_tab1}
\end{table}

\begin{table}[htbp]
    \centering
    \begin{tabular}{lrrrrrr}
    \toprule
          & Dist-1 & Dist-2 & Novelty & Quality & Diversity \\\midrule
        LM  &0.87&0.94&5.81&4.28&4.02  \\
        Retrieval &0.97 &0.87  &2.10&2.40&1.76 \\
        BERT &0.91&0.88&1.86&1.82&2.15 \\
        \bottomrule
    \end{tabular}
    \vspace{3mm}
    \caption{Evaluation results for generation-based K2S on English sentences}
    \label{tab:text_infill_tab2}
\end{table}

\section{Conclusion}

In this technical report, we have introduced Effidit, a writing assistant that facilitates users to write high quality text efficiently by using AI technologies.
Our system supports two languages (Chinese and English) and two domains (general and academic).
Five categories of functions are provided: text completion, error checking, text polishing, K2S, and cloud IME.
With these functions, Effidit significantly expands the capacities of a typical writing assistant.
We have introduced the overall system, its major functions, and the technical implementation of some key modules.
In the future, we plan to keep improving the quality of each module to make the system more helpful and easy-to-use.

\bibliographystyle{unsrt}  
\bibliography{references}

\end{CJK*}
\end{document}